\documentclass[10pt,twocolumn,letterpaper]{article}

\usepackage{cvpr}

\graphicspath{{figure/}}
\usepackage{overpic,wrapfig}

\usepackage{cuted}
\usepackage{caption}

\usepackage{mathtools}
\usepackage{siunitx}

\usepackage[linesnumbered,ruled,vlined]{algorithm2e}

\usepackage[acronym]{glossaries}
\usepackage{balance}
\usepackage{xspace}

\usepackage{soul}
\usepackage[normalem]{ulem}

\usepackage{tabularx,multirow,makecell}

\usepackage{pifont}

\definecolor{cvprblue}{rgb}{0.21,0.49,0.74}
\usepackage[pagebackref,breaklinks,colorlinks,allcolors=cvprblue]{hyperref} 
\usepackage[capitalize]{cleveref}

\newcommand{\cmark}{\ding{51}}

\definecolor{oursgray}{gray}{0.94}

\crefname{algorithm}{Alg.}{Algs.}
\Crefname{algocf}{Algorithm}{Algorithms}
\crefname{section}{Sec.}{Secs.}
\Crefname{section}{Section}{Sections}
\crefname{table}{Tab.}{Tabs.}
\Crefname{table}{Table}{Tables}
\crefname{figure}{Fig.}{Figs.}
\Crefname{figure}{Figure}{Figures}
\crefname{equation}{Eq.}{Eqs.}
\Crefname{equation}{Equation}{Equations}
\crefname{appendix}{Appx.}{Appxs.}
\Crefname{appendix}{Appendix}{Appendices}

\newcommand{\dataset}{DexHOI\xspace}

\newacronym{hoi}{HOI}{hand-object interaction}
\newacronym{iou}{IoU}{intersection over union}
\newacronym{pbr}{PBR}{physically based rendering}
\newacronym{pnp}{PnP}{perspective-n-point}

\newcommand\customparagraph[1]{\vspace{0.4em}\noindent\textbf{#1}}

\def\confName{CVPR}
\def\confYear{2027}

\title{\dataset: A Dataset of High-Speed Hand-Object Interactions\\for Understanding and Learning Dexterous Manipulation}

\author{Leiyao Cui\quad Lifang Liang\quad Lecheng Ruan\textsuperscript{\textdagger}\quad Zihang Zhao\textsuperscript{\textdagger}\\
\url{https://dexhoi27.github.io}
}

\begin{document}
\maketitle

\begingroup
\renewcommand{\thefootnote}{\fnsymbol{footnote}}
\footnotetext[0]{\mbox{All authors are with HEC Future Lab.\quad
\textsuperscript{\textdagger} Corresponding authors.}}
\endgroup

\begin{strip}
    \vspace{-40pt}

    \centering
    \captionsetup{type=figure}
    \includegraphics[width=\linewidth]{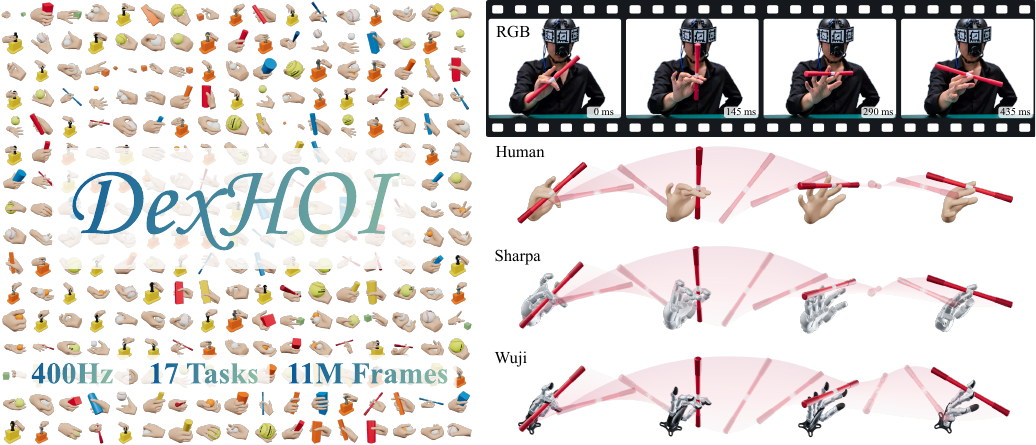}
    \caption{\textbf{Overview of \dataset.} The dataset comprises approximately 11M frames of dexterous, high-speed \acrlongpl{hoi} captured by synchronized eight cameras at 400\,Hz across 17 diverse tasks. The examples (right) illustrate a pen-spinning clip performed by a human and two robotic platforms, highlighting the potential of \dataset for transferring dexterous manipulation skills to robots.}
    \label{fig:teaser}
\end{strip}

\begin{abstract}
Capturing fine-grained human \glspl{hoi} is essential for machines to understand and learn manipulation skills. However, existing datasets offer limited coverage of dexterous manipulation involving intricate hand movements and complex object dynamics, such as pen spinning. To address this gap, we introduce \textbf{\dataset}, a high-speed, multi-view dataset of \glspl{hoi} captured by eight synchronized cameras operating at 400\,Hz. \dataset comprises 272 sequences across 17 dexterous manipulation tasks, totaling 11 million frames, and provides fine-grained observations of rapid hand movements and object dynamics. To demonstrate its broad utility, we establish three complementary benchmarks spanning perception, prediction, and robot learning: (1) egocentric hand reconstruction, which recovers hand poses from egocentric visual observations; (2) hand-conditioned object dynamics prediction, which forecasts future object states conditioned on the current object state and hand actions; and (3) dexterous robot learning, which leverages human interaction data to learn robotic dexterous manipulation policies. We envision \dataset as a foundation for advancing high-speed hand-object interaction understanding, object dynamics modeling, and human-inspired dexterous robot learning.
\end{abstract}

\section{Introduction}

Human hands can perform a wide range of dexterous manipulation skills, from handling everyday objects to executing intricate actions such as pen spinning~\cite{feix2015grasp,sobinov2021neural,zhao2025embedding,luo2026t}. For machines to understand and learn these skills, large-scale, fine-grained observations of human \glspl{hoi} are highly appreciated~\cite{huang2016recent,woo2023survey}. This need has motivated extensive efforts to collect human \gls{hoi} data~\cite{garcia2018first,zimmermann2019freihand,hasson2019learning,hampali2020honnotate,brahmbhatt2020contactpose,chao2021dexycb,kwon2021h2o,yang2022oakink,liu2022hoi4d,fan2023arctic,zhan2024oakink2,liu2024taco,wang2025ho,fu2025gigahands,banerjee2025hot3d,rim2026show3d,engelbracht2026hoi}, which have, in turn, fueled substantial research progress in motion synthesis~\cite{li2024controllable,di2026pamotion,Zhang2026handx}, generative modeling~\cite{cha2024text2hoi,zhang2024hoidiffusion,jung2026decoupled}, robot learning~\cite{qin2022dexmv,hong2026learning,cui2026vi}, and \etc.

Despite significant progress, existing \gls{hoi} datasets provide limited coverage of high-speed dexterous manipulation. In particular, skills such as pen spinning involve rapid finger articulation, frequent changes in hand-object contact, and complex object dynamics~\cite{andrychowicz2020learning,charlesworth2021solving,wang2025lessons,zhao2025tac}. This leaves high-speed human \gls{hoi} underexplored and limits the data available for studying such interactions.

To bridge this gap, we introduce \dataset, a high-speed, multi-view dataset of human \glspl{hoi}, as shown in \cref{fig:teaser}. To minimize interference with natural human manipulation, we capture interactions using only eight synchronized RGB cameras operating at 400\,Hz, comprising one egocentric and seven exocentric cameras. The dataset includes 16 subjects, 30 objects, and 272 sequences spanning 17 dexterous manipulation tasks, totaling 11 million frames. Through an automated processing pipeline, each frame provides eight RGB views, hand masks, 6D object poses, and reconstructed 3D hand geometry with a median reprojection IoU of 0.89~\cite{zimmermann2019freihand,cao2021reconstructing,Rozumnyi2023_estimating}.

To demonstrate the utility of \dataset, we establish three benchmarks spanning perception, prediction, and robot learning, with the goal of advancing the understanding and learning of dexterous manipulation. First, \textbf{egocentric hand reconstruction} focuses on recovering 3D hand geometry from egocentric visual observations. Second, \textbf{hand-conditioned object dynamics prediction} investigates how hand actions influence object motion by forecasting future object states conditioned on the current object state and hand actions. Third, \textbf{dexterous robot learning} explores how human \gls{hoi} data can be leveraged to learn dexterous manipulation policies for robots. Together, these benchmarks provide a foundation and new challenges for developing machines that can better perceive, predict, and reproduce the high-speed dexterous manipulation skills demonstrated by humans.

In summary, our main contributions are threefold:
\begin{itemize}
\item \textbf{A high-speed, multi-view dataset of dexterous \glspl{hoi}.} We introduce \dataset, comprising 272 sequences across 17 dexterous manipulation tasks and 11 million frames, providing fine-grained observations of rapid, complex hand-object dynamics.
\item \textbf{An automated data acquisition pipeline.} We develop an automated pipeline based on synchronized RGB cameras to capture high-speed \glspl{hoi} while minimizing interference with natural human manipulation.
\item \textbf{Three benchmarks for high-speed \gls{hoi}.} We benchmark three complementary tasks for advancing the understanding and learning of dexterous manipulation, highlighting the new challenges posed by \dataset.
\end{itemize}

\begin{table*}[!t]
    \centering
    \caption{\textbf{Comparison of our \dataset with the publicly available datasets of \glspl{hoi}.} *Free Hand: Indicates whether hands are free of gloves and markers. Hand Mask: Indicates whether hand masks are extracted from real images rather than rendered images.}
    \label{tab:related_work}

    \setlength{\tabcolsep}{4.8pt}
    \begin{tabular}{lcccccccccccc}
        \toprule
         \multirow{2}{*}[-3.5pt]{Dataset} & \multicolumn{5}{c}{Number of} & \multirow{2}{*}[-3.5pt]{FPS} & \multirow{2}{*}[-3.5pt]{\shortstack{Image\\Resolution}} & \multirow{2}{*}[-3.5pt]{\shortstack{Free\\Hand}} & \multirow{2}{*}[-3.5pt]{\shortstack{Hand\\Mask}} & \multirow{2}{*}[-3.5pt]{\shortstack{Hand\\Geom.}} & \multirow{2}{*}[-3.5pt]{\shortstack{Obj.\\Pose}} & \multirow{2}{*}[-3.5pt]{\shortstack{High\\Speed}}\\
          \cmidrule(lr){2-6}
          & Subj. & Obj. & Tasks & Frames & Exo\,/\,Ego & & & & & & &\\
          \midrule
          FHPA~\cite{garcia2018first} & 6 & 26 & 45 & 0.1M & 0\,/\,1 & 30 & 1920\(\times\)1080 &  & & & \cmark & \\
          FreiHAND~\cite{zimmermann2019freihand} & 32 & - & - & 37K & 8\,/\,0 & - & 224\(\times\)224  & \cmark & \cmark & \cmark & & \\
          HO-3D~\cite{hampali2020honnotate} & 10 & 10 & - & 78k & 5\,/\,0 & - & 640\(\times\)480 & \cmark & & \cmark & \cmark & \\
          ContactPose~\cite{brahmbhatt2020contactpose} & 50 & 25 & 2 & 2.9M & 3\,/\,0 & 30 & 1920\(\times\)1080 & \cmark & \cmark & \cmark & \cmark & \\
          DexYCB~\cite{chao2021dexycb} & 10 & 20 & 1 & 0.6M & 8\,/\,0 & 30 & 640\(\times\)480 & \cmark & & \cmark & \cmark & \\
          H2O~\cite{kwon2021h2o} & 4 & 8 & 36 & 0.6M & 4\,/\,1 & 30 & 1280\(\times\)720 & \cmark & & \cmark & \cmark & \\
          OakInk~\cite{yang2022oakink} & 12 & 100 & 5 & 0.2M & 4\,/\,0 & - & 840\(\times\)480 & \cmark & & \cmark & \cmark & \\
          HOI4D~\cite{liu2022hoi4d} & 9 & 800 & 54 & 2.4M & 0\,/\,1 & 15 & 1280\(\times\)720 & \cmark & \cmark & \cmark & \cmark & \\
          ARCTIC~\cite{fan2023arctic} & 10 & 11 & 2 & 2.1M & 8\,/\,1 & 30 & 2800\(\times\)2000 & & & \cmark & \cmark & \\
          OakInk2~\cite{zhan2024oakink2} & 9 & 75 & 4 & 4.0M & 3\,/\,1 & 30 & 848\(\times\)480 & & & \cmark & \cmark & \\
          TACO~\cite{liu2024taco} & 14 & 196 & 15 & 5.2M & 12\,/\,1 & 30 & 4096\(\times\)3000 & \cmark & \cmark & \cmark & \cmark & \\
          Ho-Cap~\cite{wang2025ho} & 9 & 64 & 3 & 0.7M & 8\,/\,1 & - & 640\(\times\)480 & \cmark & \cmark & \cmark & \cmark & \\
          GigaHands~\cite{fu2025gigahands} & 56 & 417 & 191 & 183M & 51\,/\,0 & 30 & 1280\(\times\)720 & \cmark & \cmark & \cmark & \cmark & \\
          HOT3D~\cite{banerjee2025hot3d} & 19 & 33 & 4 & 3.7M & 0\,/\,3 & 30 & 1408\(\times\)1408 & & & \cmark & \cmark & \\
          Show3D~\cite{rim2026show3d} & 30 & 21 & - & 4.3M & 8\,/\,2 & 60 & 1024\(\times\)1280 & \cmark & \cmark & \cmark & \cmark & \\
          Hoi!~\cite{engelbracht2026hoi} & 7 & 381 & - & - & 2\,/\,1 & 60 & - & \cmark &  & \cmark & \cmark & \\
          \midrule
         \dataset (ours) & 16 & 30 & 17 & 10.6M & 7\,/\,1 & 400 & 1920\(\times\)1440 & \cmark & \cmark & \cmark & \cmark & \cmark \\
         \bottomrule
    \end{tabular}
\end{table*}

\section{Related Work}

\subsection{HOI Datasets}

As shown in \cref{tab:related_work}, substantial efforts have been devoted to collecting \gls{hoi} data. Early datasets, such as FreiHAND~\cite{zimmermann2019freihand} and HO-3D~\cite{hampali2020honnotate}, provide hand pose and shape annotations for hand reconstruction, while ContactPose~\cite{brahmbhatt2020contactpose} and DexYCB~\cite{chao2021dexycb} further incorporate object geometry and interaction information to study hand-object contact and grasping. FHPA~\cite{garcia2018first}, H2O~\cite{kwon2021h2o}, and HOT3D~\cite{banerjee2025hot3d} extend data collection to egocentric \glspl{hoi}. Other datasets target more diverse and dexterous manipulation scenarios. OakInk~\cite{yang2022oakink} provides a large-scale knowledge repository of \glspl{hoi}, while ARCTIC~\cite{fan2023arctic} and OakInk2~\cite{zhan2024oakink2} focus on bimanual manipulation and complex task completion. TACO~\cite{liu2024taco} emphasizes generalizable bimanual tool-action-object understanding, while GigaHands~\cite{fu2025gigahands} substantially scales up interaction diversity with 183 million frames spanning 191 tasks. More recent datasets, including Ho-Cap~\cite{wang2025ho}, Show3D~\cite{rim2026show3d}, and Hoi!~\cite{engelbracht2026hoi}, further advance 3D reconstruction, pose tracking, and multimodal understanding of \glspl{hoi}. Complementing these efforts, \dataset captures high-speed \gls{hoi} data for advancing the understanding of dexterous manipulation and facilitating the learning of dexterous manipulation policies.

\subsection{Egocentric Hand Reconstruction}

Accurate single-view hand reconstruction is a fundamental capability for machine understanding of \gls{hoi}, as it requires recovering complete 3D hand geometry from limited visual observations while resolving inherent depth ambiguities. Frame-wise approaches have explored a range of reconstruction strategies, including parametric hand model regression~\cite{boukhayma20193d, zhang2019end}, differentiable rendering~\cite{baek2019pushing}, and graph- or mesh-based prediction~\cite{ge20193d, kulon2020weakly}. More recent methods enhance robustness to occlusions and in-the-wild variations through attention-based architectures, including HandOccNet~\cite{park2022handoccnet}, HaMeR~\cite{pavlakos2024reconstructing}, and WiLoR~\cite{potamias2025wilor}. Beyond frame-wise estimation, temporal approaches leverage video information to improve reconstruction accuracy and temporal consistency~\cite{tekin2019h+,chen2021temporal,ziani2022tempclr,tu2023consistent}. More recently, Deformer~\cite{fu2023deformer} and HMP~\cite{duran2024hmp} leverage learned hand motion priors to improve reconstruction robustness under occlusion and motion ambiguity. HaWoR~\cite{zhang2025hawor} further extends this line of research to world-space hand motion estimation by explicitly accounting for egocentric camera movement. In this context, \dataset provides valuable supervision for advancing single-view hand reconstruction and \gls{hoi} understanding.

\subsection{Dynamics Prediction}

A complementary task for understanding \gls{hoi} data is forecasting how interactions influence an object's future state and is closely related to the concept of world models. Early works learn forward dynamics from physical interactions and videos through pixel-level prediction~\cite{finn2016unsupervised}, abstract feature representations~\cite{agrawal2016learning}, rigid-body transformations~\cite{byravan2017se3}, and implicit physics engines~\cite{watters2017visual,sanchez2018graph}. More recent approaches increasingly adopt object-centric world models for robotic manipulation~\cite{zhu2025irasim,zhao2026tacman}, with emerging efforts extending dynamics prediction to dexterous \glspl{hoi} through human demonstrations and explicit physical-state representations~\cite{kim2026dexterous,goswami2026world,lee2026dexplicit,hong2026learning}. \dataset supports this research direction by providing diverse, high-speed \gls{hoi} sequences with fine-grained 3D hand annotations and 6D object poses, enabling temporally dense supervision for learning action-conditioned object dynamics.

\subsection{Dexterous Robot Learning}

Learning from human demonstrations offers a promising avenue for acquiring dexterous manipulation skills, as human motions provide rich examples of coordinated finger movements and contact-rich interactions. Early works integrate human demonstrations with reinforcement learning~\cite{gupta2016learning, ruppel2020learning}, while vision-based teleoperation enables the collection of human hand motions for robotic execution~\cite{handa2020dexpilot}. Subsequent studies explore learning manipulation skills from human videos and transferring them across different hand embodiments~\cite{qin2022dexmv,qin2022one,mandikal2022dexvip,shaw2024learning}. More recent approaches advance demonstration acquisition and human-to-robot skill transfer through scalable motion capture~\cite{wang2024dexcap}, simulated demonstration augmentation~\cite{wang2024cyberdemo}, and improved imitation learning and policy transfer strategies~\cite{kim2024goal, lum2025crossing, li2025maniptrans}. Building on these developments, recent work leverages human \gls{hoi} data to learn object-centric motion priors for robotic manipulation~\cite{hong2026learning}. \dataset complements these efforts by providing diverse, high-speed \gls{hoi} sequences with fine-grained 3D hand geometry and 6D object poses, offering a rich source for human-inspired dexterous learning.

\section{The \dataset Dataset}

\dataset is a large-scale dataset of high-speed dexterous \glspl{hoi}, comprising 10.6M frames from 272 sequences spanning 17 manipulation tasks, with 16 subjects interacting with 30 objects (see \cref{tab:related_work}). To capture rapid hand movements and complex object dynamics while minimizing interference with natural human manipulation, we record these interactions using eight synchronized RGB cameras, comprising seven exocentric cameras and one egocentric camera, operating at 400\,Hz. Each camera captures images at a resolution of 1920$\times$1440 pixels. This high-speed, multi-view acquisition setup provides temporally dense observations of dexterous interactions, substantially complementing existing \gls{hoi} datasets.

\customparagraph{High-speed dexterous manipulation tasks.}
To capture the diverse behaviors and interaction dynamics involved in dexterous manipulation, \dataset covers 17 tasks organized into six categories: (1) five toss-and-catch tasks involving balls of varying sizes and weights; (2) two in-hand rotation tasks involving pairs of balls with different frictional properties; (3) four object-rotation tasks involving prisms and cylinders, each in two sizes; (4) four nut-twisting tasks involving two types of nut-and-bolt combinations with different frictional properties, each performed with two grasp configurations; (5) sponge-block flicking; and (6) diverse pen-spinning actions. Together, these tasks encompass a broad spectrum of manipulation behaviors, from object throwing and catching to continuous in-hand reorientation and intricate finger-driven object motions.

\begin{figure}
    \centering
    \includegraphics[width=\linewidth]{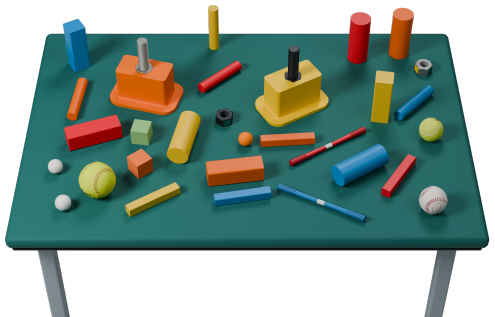}
    \caption{\textbf{Diverse 3D object models in \dataset.} The dataset provides 30 objects with high-fidelity geometry and high-quality \acrshort{pbr} materials.}
    \label{fig:object}
\end{figure}

\customparagraph{High-quality annotations.}
\dataset provides comprehensive frame-level annotations of \glspl{hoi}, including 3D hand poses and meshes represented in both MHR~\cite{ferguson2025mhr} and MANO~\cite{romero2017embodied} parameterizations, 6D object poses, and hand segmentation masks. For hand annotations, we reconstruct 3D hand meshes from synchronized multi-view RGB observations using SilHand~\cite{cui2026hand}. We further apply optimization-based temporal refinement to improve the smoothness and consistency of the reconstructed hand motions. To evaluate reconstruction quality, we project the reconstructed hand meshes into all eight camera views and compare the resulting masks against the corresponding Sapiens2~\cite{khirodkar2026sapiens2}-predicted segmentation masks, achieving a high median reprojection IoU of 0.89. For object annotations, we adopt the multi-view model-based object pose estimation pipeline introduced in SHOW3D~\cite{rim2026show3d}, which has been validated against MoCap-derived references. We further apply temporal refinement to the estimated object poses to improve their consistency across frames. To support RGB-based object pose estimation, we provide 30 textured 3D object models with high-fidelity geometry and physically based rendering (PBR) materials (\cref{fig:object}), which serve as geometric and appearance references for model-based estimation while enabling photorealistic rendering.

\begin{figure}[!t]
    \centering
    \includegraphics[width=\linewidth]{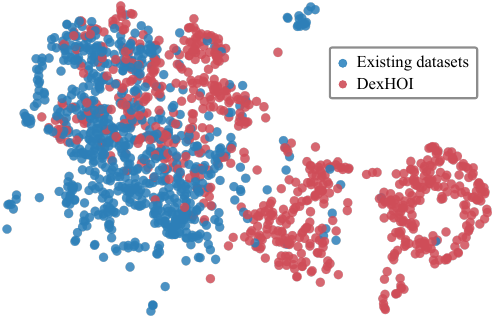}
    \caption{\textbf{Contact pattern comparison between \dataset and existing datasets.} UMAP visualization of frame-wise hand-object contact patterns, where each point represents an individual contact pattern. Red and blue points denote patterns from \dataset and the eleven existing datasets, respectively.}
    \label{fig:contact}
\end{figure}

\customparagraph{Diverse hand-object contact patterns.}
Beyond task and object diversity, \dataset captures a broad range of hand-object contact configurations arising from dexterous manipulation. These configurations are particularly important for understanding fine-grained interactions involving frequent changes in contact during rapid hand movements. Following GRAB~\cite{taheri2020grab}, we derive frame-wise, vertex-level binary contact labels from the reconstructed hand and object meshes, accounting for both mesh interpenetration and proximity-based contact. We use these labels to characterize the contact patterns in \dataset and compare their distributions with those from 11 existing datasets~\cite{hampali2020honnotate,chao2021dexycb,kwon2021h2o,yang2022oakink,fan2023arctic,zhan2024oakink2,liu2024taco,fu2025gigahands,banerjee2025hot3d,wang2025ho,rim2026show3d} through a UMAP~\cite{mcinnes2018umap} embedding. As shown in \cref{fig:contact}, contact patterns from \dataset occupy distinct regions of the embedding space, suggesting that \dataset captures complementary contact configurations that are less represented in existing datasets. This broader contact coverage, together with the dataset's high temporal resolution and diverse manipulation tasks, makes \dataset a valuable resource for advancing fine-grained contact modeling and understanding high-speed dexterous manipulation.

\section{Data Collection and Annotation}

\begin{figure*}[!t]
    \centering
    \includegraphics[width=\linewidth]{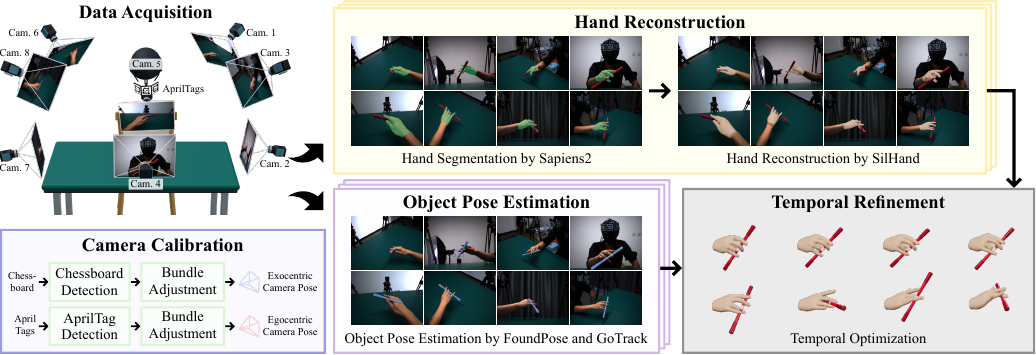}
    \caption{\textbf{Data collection and annotation pipeline.} The \dataset captures synchronized multi-view RGB observations using seven exocentric cameras and one headset-mounted egocentric camera. We calibrate the exocentric camera rig using chessboard observations and estimate frame-wise egocentric camera poses using headset-mounted AprilTags. We then reconstruct 3D hand geometries from Sapiens2-predicted segmentation masks using SilHand, while estimating object poses using FoundPose and refining them with GoTrack. Finally, we apply optimization-based temporal refinement to improve the temporal consistency of the hand and object trajectories.}
    \label{fig:pipeline}
\end{figure*}

We develop a data collection and annotation pipeline to obtain high-quality 3D annotations of high-speed \glspl{hoi} from synchronized multi-view RGB observations. As illustrated in \cref{fig:pipeline}, the pipeline consists of five stages: data acquisition (\cref{sec:data_acquisition}), camera calibration (\cref{sec:camera_calibration}), hand reconstruction (\cref{sec:hand_reconstruction}), object pose estimation (\cref{sec:object_pose_estimation}), and temporal refinement (\cref{sec:temporal_refinement}). The first two stages establish synchronized visual observations and a consistent geometric reference, upon which we perform hand reconstruction and object pose estimation. Finally, we jointly refine the resulting hand and object trajectories to improve their temporal consistency.

\subsection{Data Acquisition System}\label{sec:data_acquisition}

We construct an eight-camera RGB capture system to record high-speed dexterous \glspl{hoi} from complementary viewpoints. As illustrated in \cref{fig:pipeline}, seven exocentric cameras (Cam. 1--4 and Cam. 6--8) are distributed around the tabletop manipulation workspace, providing multi-view observations of the subject's hands and manipulated objects. To capture first-person observations, we mount an additional egocentric camera (Cam. 5) on a custom-built headset worn by the subject. We further attach three AprilTags~\cite{krogius2019flexible} to the headset to facilitate frame-wise estimation of the egocentric camera pose. All eight cameras are synchronized and capture RGB images at 400\,Hz with a resolution of 1920$\times$1440 pixels. By relying solely on RGB cameras, our acquisition system minimizes interference with natural human manipulation while providing temporally dense visual observations of rapid hand movements and complex object dynamics.

\subsection{Camera Calibration}\label{sec:camera_calibration}

To establish a consistent geometric reference for multi-view reconstruction, we estimate the extrinsic poses of all eight cameras. For the seven fixed exocentric cameras, we calibrate the camera rig separately for each subject using synchronized chessboard observations collected in a dedicated calibration sequence. We jointly optimize the camera poses and per-frame chessboard poses through bundle adjustment, minimizing the reprojection errors.

Unlike fixed exocentric cameras, the egocentric camera moves with the subject and therefore requires frame-wise pose estimation. To this end, we adopt an AprilTag-based approach. We first initialize the rigid transformations between the three headset-mounted AprilTags and the egocentric camera using a hand-eye calibration method~\cite{shah2013solving} on a dedicated calibration recording, and subsequently refine these transformations through bundle adjustment. During each manipulation sequence, we detect the headset-mounted AprilTags in the exocentric views and use the calibrated transformations to recover the frame-wise egocentric camera poses. We further jointly optimize these poses over each sequence using robust tag reprojection errors, together with linear and angular jerk regularization, to obtain a temporally consistent egocentric camera trajectory.

\subsection{Hand Reconstruction}\label{sec:hand_reconstruction}

Given the calibrated multi-view observations, we reconstruct 3D hand geometries using SilHand~\cite{cui2026hand}. Specifically, we first employ Sapiens2~\cite{khirodkar2026sapiens2} to predict hand segmentation masks from the captured RGB images and manually refine them. Using the synchronized multi-view hand masks and calibrated camera parameters, SilHand reconstructs the 3D hand geometry in the MHR~\cite{ferguson2025mhr} parameterization with the global hand pose.

To maintain consistent hand geometry across sequences, we first estimate a subject-specific hand shape from a dedicated recording sequence and reuse it throughout all manipulation sequences performed by the same subject. The resulting hand reconstructions are subsequently refined using multi-view geometric constraints and temporal optimization, as described in \cref{sec:temporal_refinement}. Finally, to support downstream tasks requiring different hand parameterizations, we convert the refined MHR-based hand reconstructions to MANO~\cite{romero2017embodied} by fitting MANO meshes to the reconstructed hand surfaces.

\subsection{Object Pose Estimation}\label{sec:object_pose_estimation}

In parallel with hand reconstruction, we estimate the 6D poses of manipulated objects using the multi-view model-based pipeline introduced in SHOW3D~\cite{rim2026show3d}. Given the textured 3D object models, CNOS~\cite{nguyen2023cnos} is first deployed for object detection and FoundPose~\cite{ornek2024foundpose} for initial 6D pose estimation. GoTrack~\cite{nguyen2025gotrack} is then used to refine the initial pose estimates by leveraging multi-view visual correspondences and geometric consistency. The resulting frame-wise object poses are further refined through temporal optimization, as described in \cref{sec:temporal_refinement}.

\subsection{Temporal Refinement}\label{sec:temporal_refinement}

Although multi-view observations provide complementary geometric constraints, independently reconstructed hand poses and estimated object poses may still exhibit temporal inconsistencies, particularly during rapid movements and occlusions. To address this issue, we perform optimization-based temporal refinement on both hand and object trajectories. Our objective is to improve temporal smoothness and consistency while aligning with RGB observations.

For hand reconstructions, we jointly refine the global hand poses and hand articulation parameters over each sequence. The optimization incorporates multi-view mask alignment, 2D keypoint reprojection errors, and priors on the initial hand estimates. To encourage temporally coherent hand motion, we further regularize the linear and angular jerk of the global hand poses, as well as the angular jerk of individual finger joints.

For object poses, we optimize the estimated 6D trajectories by balancing robust, symmetry-aware pose consistency terms with temporal regularization on linear and angular jerk. This refinement reduces temporal fluctuations while preserving consistency with the initial pose estimates and accounting for the geometric symmetries of the manipulated objects.

\section{Experiments}

To demonstrate the utility of \dataset, we establish and evaluate three complementary benchmarks spanning perception, prediction, and robot learning: egocentric hand reconstruction (\cref{sec:egocentric_hand_reconstruction}), hand-conditioned object dynamics prediction (\cref{sec:object_dynamics_prediction}), and dexterous robot learning (\cref{sec:dexterous_robot_learning}). Together, these benchmarks evaluate the potential of high-speed human \glspl{hoi} data to reconstruct hand geometry, model object dynamics, and learn dexterous manipulation skills.

\begin{table}[!t]
    \centering
    \caption{\textbf{Evaluation of egocentric hand reconstruction.} All metrics are reported in millimeters ($\downarrow$: lower is better), with the best result highlighted in \textbf{bold}. The superscript * denotes models fine-tuned on \dataset.}
    \label{tab:benchmark_hand_reconstruction}

    \small
    \begin{tabular}{lcccc}
        \toprule
        Method & MPVPE $\downarrow$ & MPJPE $\downarrow$ & PA-V $\downarrow$ & PA-J $\downarrow$ \\
        \midrule
        HaMeR~\cite{pavlakos2024reconstructing} & 27.34 & 28.46 & 8.47 & 8.31 \\
        HaMeR*~\cite{pavlakos2024reconstructing}  & \textbf{11.07} & \textbf{11.08} & \textbf{5.23} & \textbf{4.98} \\
        \midrule
        HaWoR~\cite{zhang2025hawor} & 22.64 & 23.67 & 9.00 & 8.91 \\
        HaWoR*~\cite{zhang2025hawor} & \textbf{9.86} & \textbf{9.89} & \textbf{5.61} & \textbf{5.45} \\
        \bottomrule
    \end{tabular}
\end{table}

\subsection{Egocentric Hand Reconstruction}\label{sec:egocentric_hand_reconstruction}

We evaluate existing methods for reconstructing 3D hand geometry from egocentric RGB observations, which is challenging in dexterous manipulation due to rapid finger articulation, frequent hand-object occlusions, and substantial variations in hand configurations.

\customparagraph{Benchmark setup.}
We evaluate HaMeR~\cite{pavlakos2024reconstructing} and HaWoR~\cite{zhang2025hawor}, considering both their original pretrained models and models fine-tuned on \dataset. HaMeR reconstructs hand geometry from individual RGB images, whereas HaWoR additionally exploits temporal information from consecutive frames. We adopt a cross-subject protocol, using 14 subjects for training and two held-out subjects for testing. All experiments use egocentric RGB observations sampled at 30\,Hz, with the corresponding 3D hand annotations serving as supervision and evaluation references.

\customparagraph{Evaluation metrics.}
We measure the reconstruction accuracy using the mean per-vertex position error (MPVPE) and mean per-joint position error (MPJPE), together with their Procrustes-aligned counterparts (PA-V and PA-J). All errors are reported in millimeters, with lower values indicating better accuracy.

\customparagraph{Results.}
As reported in \cref{tab:benchmark_hand_reconstruction}, fine-tuning on \dataset consistently improves both methods across all metrics. For HaMeR, MPVPE and MPJPE decrease from 27.34\,mm and 28.46\,mm to 11.07\,mm and 11.08\,mm, respectively. HaWoR achieves further improvements, reducing these errors from 22.64\,mm and 23.67\,mm to 9.86\,mm and 9.89\,mm. In addition, fine-tuned HaWoR achieves the lowest unaligned errors, while fine-tuned HaMeR performs slightly better under Procrustes alignment. These results demonstrate that \dataset provides valuable supervision for adapting existing methods to high-speed \glspl{hoi} data, while the remaining errors highlight the challenges of recovering hand geometry from egocentric observations in dexterous interactions.

\begin{table}[!t]
    \centering
    \caption{\textbf{Evaluation of object dynamics prediction.} Translation and rotation errors are reported in millimeters and degrees, respectively. $\downarrow$ indicates that lower values are better. The best result for each metric is highlighted in \textbf{bold}. The superscript * denotes models fine-tuned on \dataset.}
    \label{tab:benchmark_object_dynamics_prediction}

    \small
    \setlength{\tabcolsep}{5.4pt}
    \begin{tabular}{lccccc}
        \toprule
        \multirow{2}{*}{Method} & Freq. & ATE $\downarrow$ & ARE $\downarrow$ & RTE $\downarrow$ & RRE $\downarrow$ \\
        & (Hz) & (mm) & ($^\circ$) & (mm) & ($^\circ$) \\
        \midrule
        DeXplicit~\cite{lee2026dexplicit} & 15 & 68.75 & 88.86 & 22.79 & \textbf{52.24} \\
        DeXplicit*~\cite{lee2026dexplicit} & 15 & \textbf{45.82} & \textbf{84.30} & \textbf{22.11} & 60.62 \\
        World Model & 15 & 59.45 & 99.32 & 36.12 & 116.96 \\
        \midrule
        World Model & 30 & 36.95 & 91.66 & 12.38 & 35.21 \\
        World Model & 60 & 31.77 & 75.32 & 5.66 & 17.89 \\
        World Model & 120 & 26.88 & 68.59 & 2.71 & 10.54 \\
        World Model & 200 & 22.13 & 50.10 & 1.50 & 4.31 \\
        World Model & 400 & 13.93 & 26.13 & 0.75 & 1.33 \\
        \bottomrule
    \end{tabular}
\end{table}

\subsection{Object Dynamics Prediction}\label{sec:object_dynamics_prediction}

We investigate hand-conditioned object dynamics prediction, which aims to forecast future object motion given its initial state and a known hand-motion trajectory. Formally, given an initial object pose $\boldsymbol{T}^{o}_{t}$ and a hand-motion sequence $\mathcal{H}_{t+1:t+K}$, the objective is to predict the future object poses:
\begin{equation}
    \hat{\boldsymbol{T}}^{o}_{t+1:t+K} = f\left(\boldsymbol{T}^{o}_{t},\mathcal{H}_{t+1:t+K}\right).
\end{equation}
Unlike conventional object motion prediction, which primarily relies on the object's motion history, this formulation explicitly conditions future object motion on hand actions. It therefore investigates how fine-grained hand articulation influences object dynamics. We focus on pen spinning, a representative high-speed manipulation task characterized by rapid finger movements and complex object dynamics.

\customparagraph{Benchmark setup.}
We evaluate DeXplicit~\cite{lee2026dexplicit} and a SuperMimic~\cite{liao2026supermimic}-inspired world model. DeXplicit takes an initial 3D scene representation and the corresponding hand-motion trajectory as input. We evaluate both its original pretrained model and the model fine-tuned on \dataset. Our world model instead directly operates on geometric hand representations, autoregressively predicting object motion from the initial object pose and consecutive MANO hand states. We adopt the same cross-subject split as the hand reconstruction benchmark, using 14 subjects for training and two held-out subjects for testing.

We first compare the two approaches at 15\,Hz using a 34-step prediction horizon, following the setting adopted in DeXplicit~\cite{lee2026dexplicit}. We further train and evaluate the world model at 30, 60, 120, 200, and 400\,Hz. This frequency sweep allows us to examine how the temporal resolution influences the prediction of rapid object dynamics.

\customparagraph{Evaluation metrics.}
We evaluate the predicted object trajectories using absolute translation error (ATE), absolute rotation error (ARE), relative translation error (RTE), and relative rotation error (RRE). ATE and ARE measure the discrepancies between predicted and ground-truth object poses over the future trajectory, whereas RTE and RRE measure the errors in relative transformations between consecutive poses. To account for the axial symmetry of the pen, we align symmetry-equivalent orientations when computing rotational errors. Translation and rotation errors are reported in millimeters and degrees, respectively, with lower values indicating better predictions.

\begin{figure}[!t]
    \centering
    \includegraphics[width=\linewidth]{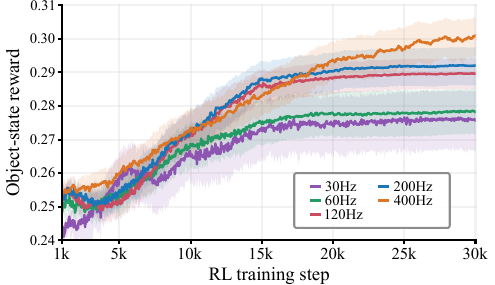}
    \caption{\textbf{Effect of reference sampling frequency on dexterous robot learning.} Object-state rewards over 30k RL training steps using demonstrations sampled at different frequencies.}
    \label{fig:quantitative_robot}
\end{figure}

\begin{figure*}[!t]
    \centering
    \includegraphics[width=\linewidth]{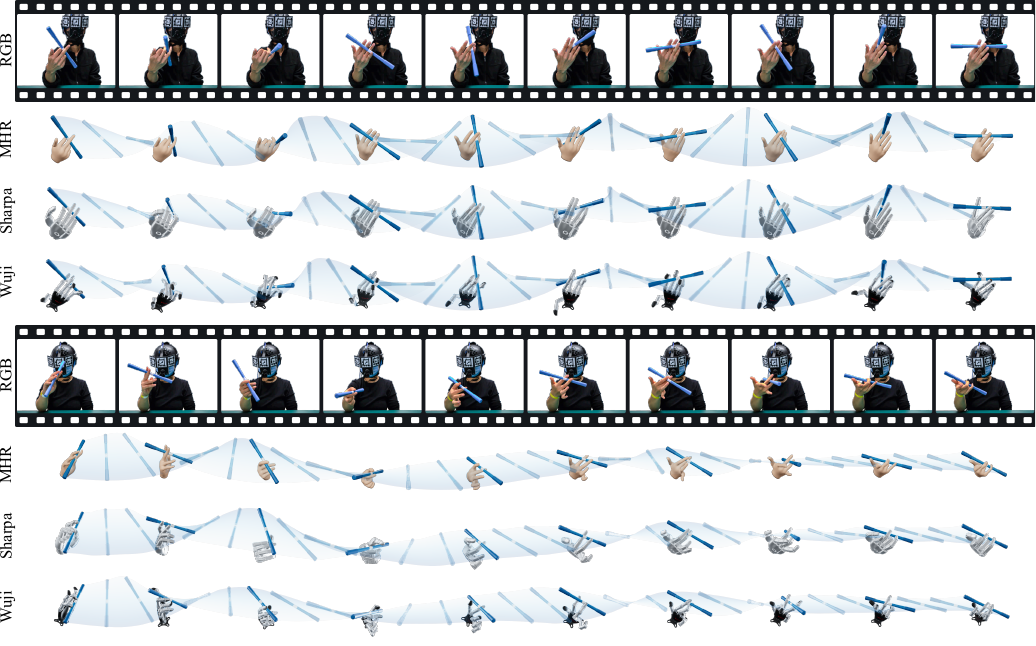}
    \caption{\textbf{Qualitative results of dexterous robot learning.} We exemplify two clips of pen-spinning skill reproduction, comparing 400\,Hz human demonstrations (RGB and MHR) with the corresponding policy executions on Sharpa Wave W01 and Wuji Hand 2.}
    \label{fig:qualitative_robot}
\end{figure*}

\customparagraph{Results.}
As reported in \cref{tab:benchmark_object_dynamics_prediction}, fine-tuning DeXplicit improves absolute object pose prediction, although accurately capturing relative rotational motion remains challenging. The frequency ablation further reveals substantial improvements in prediction accuracy as the temporal sampling rate increases. Importantly, we maintain a fixed 34-step prediction horizon across all sampling rates, meaning that higher sampling frequencies correspond to shorter physical prediction horizons. This setting is particularly relevant to high-speed robotic manipulation, where accurate short-horizon predictions are essential for frequent action updates and responsive motion control. Rather than relying on long-term predictions, which may become increasingly unreliable as manipulation states and contact configurations evolve, a robot can repeatedly predict near-future object motion, refine its actions based on new observations, and adapt to rapid changes in the interaction. Consequently, accurate short-horizon object dynamics prediction is especially valuable in this setting. Our results demonstrate that \dataset's temporally dense observations can support the learning of fine-grained, short-horizon object dynamics prediction, providing a foundation for robots to develop the predictive capabilities required for high-speed dexterous manipulation.

\subsection{Dexterous Robot Learning}\label{sec:dexterous_robot_learning}

We further investigate the utility of \dataset for learning dexterous manipulation policies on robotic platforms. Based on the aforementioned world model, we train sequence-specific reference-tracking policies to reproduce human pen-spinning motions on two robotic hands: Wuji Hand 2 and Sharpa Wave W01. Specifically, each policy is trained using an individual pen-spinning sequence from \dataset as its reference demonstration.

\customparagraph{Experiment setup.}
We conduct all experiments in MuJoCo Warp~\cite{mujoco_warp} with a simulation frequency of 2000\,Hz. For each robotic hand and reference sequence, we train a separate policy for 30k training steps. To investigate the effect of reference temporal resolution on policy learning, we sample the human demonstrations at 30, 60, 120, 200, and 400\,Hz and compare the resulting object-state reward curves. The object-state reward evaluates the agreement between the executed and reference pen poses, combining the Euclidean translation error with the Frobenius norm of the rotational discrepancy after accounting for the pen's symmetry.

\customparagraph{Results.}
As shown in \cref{fig:quantitative_robot}, policies trained with demonstrations sampled at 400\,Hz achieve the highest object-state reward. Although lower-frequency demonstrations lead to faster initial learning, their performance gradually plateaus. The superior performance at 400\,Hz suggests that fine-grained temporal information, which may be lost at lower sampling frequencies, can be effectively exploited by RL to learn more precise manipulation behaviors.

Finally, we report the tracking success rates of policies trained on 400\,Hz demonstrations. We define the tracking success rate as the fraction of the total reference duration during which the pen does not fall from the hand. Under this metric, Wuji Hand 2 and Sharpa Wave W01 achieve tracking success rates of 92.36\,\% and 90.52\,\%, respectively, demonstrating the effectiveness of the learned policies across different robotic hands. Qualitative comparisons are presented in \cref{fig:qualitative_robot}.

\section{Conclusion}

We introduce \dataset, a high-speed, multi-view dataset of dexterous \glspl{hoi}, comprising 272 sequences across 17 tasks, 16 subjects, and 30 objects, totaling 11M frames at 400\,Hz. Featuring synchronized egocentric and exocentric views, comprehensive 3D annotations, and diverse contact patterns, \dataset captures rapid hand articulation and complex object dynamics. We further establish three benchmarks for egocentric hand reconstruction, hand-conditioned object dynamics prediction, and dexterous robot learning, demonstrating its utility for perception, prediction, and robotic manipulation. We hope \dataset will advance research in high-speed dexterous manipulation.

\clearpage

\section*{LLM Usage Statement}

In this work, we used ChatGPT for language polishing and code review.

{
    \small
    \bibliographystyle{ieeenat_fullname}
    \bibliography{reference_header,reference}
}

\clearpage
\setcounter{page}{1}
\maketitlesupplementary

\section{Additional Dataset Statistics}

\begin{figure}[!t]
    \centering
    \includegraphics[width=\linewidth]{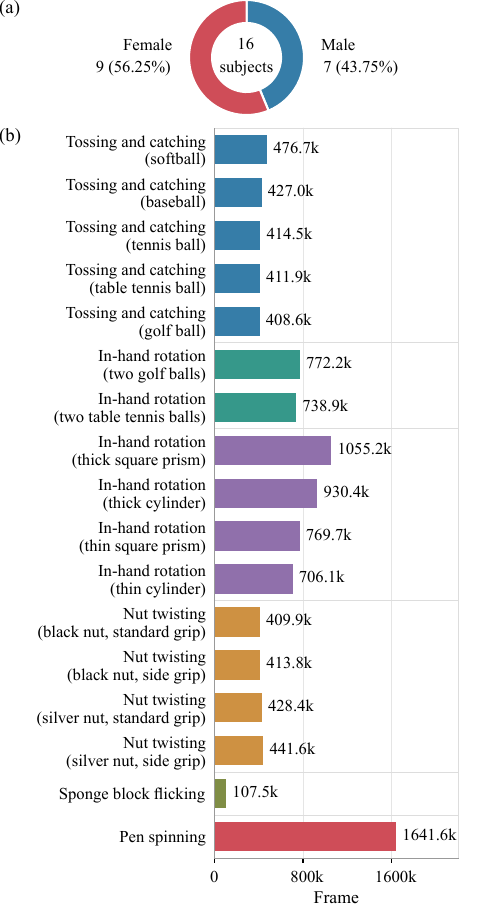}
    \caption{\textbf{The \dataset dataset statistics.} (a) Gender distribution of the 16 subjects, comprising 7 males and 9 females. (b) Distribution of 10.6M frames over 17 dexterous manipulation tasks.}
    \label{app:fig:statistics}
\end{figure}

We recruited 16 subjects with diverse hand shapes, including 7 males and 9 females. All subjects provided written informed consent for the collection and use of their data for research purposes, and received monetary compensation for their participation. \Cref{app:fig:statistics} summarizes the dataset statistics, showing the subject demographics in (a) and the distribution of image frames across the 17 manipulation tasks in (b).

\section{Ethical Approval}

This work involved human subjects. All data collection procedures and experimental protocols were approved by the relevant Institutional Review Board (IRB). Details of the approving IRB will be disclosed upon acceptance to preserve anonymity during peer review.

\begin{figure}[!t]
    \centering
    \includegraphics[width=\linewidth]{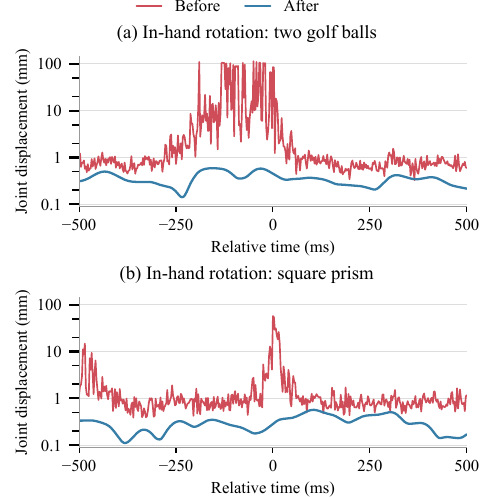}
    \caption{\textbf{Effect of temporal refinement on hand reconstruction.} Mean inter-frame 3D displacement before and after refinement for (a) rotation of two golf balls and (b) a square prism. Each plot spans one second at 400\,Hz, with displacement shown in millimeters on a logarithmic scale.}
    \label{app:fig:hand_refinement}
\end{figure}

\section{Temporal Refinement Details}

\begin{figure*}[!t]
    \centering
    \includegraphics[width=\linewidth]{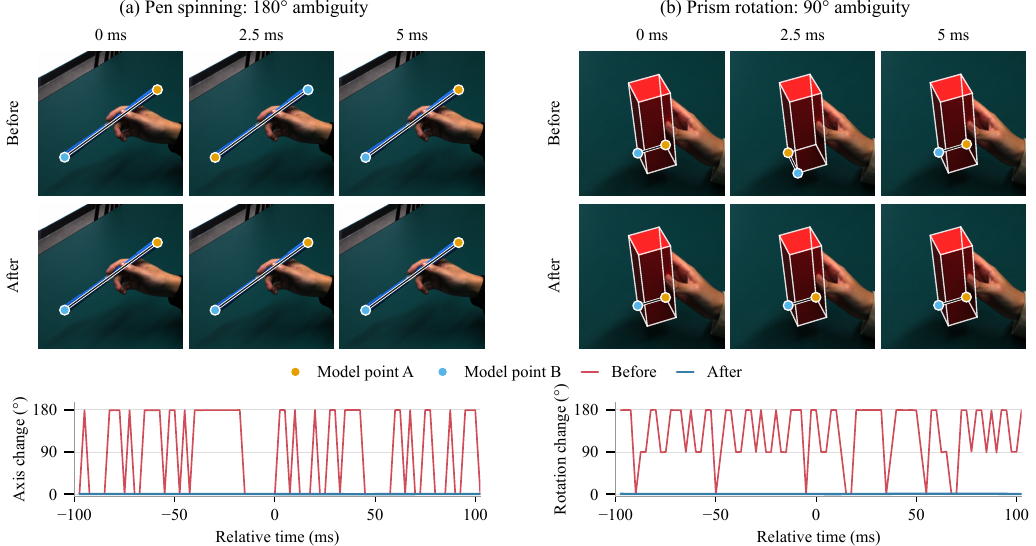}
    \caption{\textbf{Symmetry-aware temporal refinement of object poses.} (a) Pen spinning exhibits a 180$^\circ$ orientation ambiguity, while (b) prism rotation exhibits a 90$^\circ$ rotational ambiguity. Symmetry-equivalent pose estimates can introduce abrupt orientation changes between consecutive frames despite similar visual appearances. The upper rows compare object poses before and after refinement, with colored points indicating fixed locations on the object models. The lower plots show the corresponding angular changes over time, illustrating how symmetry-aware refinement suppresses artificial orientation discontinuities.}
    \label{app:fig:symmetry}
\end{figure*}

We provide additional details of the optimization-based temporal refinement introduced in \cref{sec:temporal_refinement}. Although multi-view reconstruction provides geometric constraints for hand reconstruction and object estimation, the resulting frame-wise annotations may contain outliers and temporal discontinuities, particularly under rapid motion, occlusion, and object symmetries. We therefore refine the hand and object trajectories at the sequence level, balancing consistency with the initial estimates and visual observations against temporal regularization.

\subsection{Hand Trajectory Refinement}

For hand reconstruction, we optimize the global hand poses and hand articulation parameters while keeping the subject-specific hand shape fixed. The optimization objective is formulated as
\begin{equation}
    \mathcal{L}_{\text{hand}} = \lambda_{\text{mask}}\mathcal{L}_{\text{mask}} +\lambda_{\text{kp}}\mathcal{L}_{\text{kp}} +\lambda_{\text{prior}}\mathcal{L}_{\text{prior}} +\lambda_{\text{temp}}\mathcal{L}_{\text{temp}},
\end{equation}
where $\mathcal{L}_{\text{mask}}$ measures the discrepancy between rendered hand masks and the corresponding segmentation masks across calibrated camera views. The keypoint term $\mathcal{L}_{\text{kp}}$ penalizes the reprojection errors of 3D hand joints against Sapiens2-predicted 2D keypoints, weighted by their detection confidence. The prior term $\mathcal{L}_{\text{prior}}$ employs the Geman--McClure robust loss to regularize deviations of the optimized global hand poses and joint rotations from the initial SilHand reconstructions.

To enhance temporal consistency, we regularize the linear and angular jerk of the global hand motion, together with the angular jerk of individual finger joints. Let $\boldsymbol{T}_t \in \text{SE}(3)$ and $\boldsymbol{R}_{t,j} \in \text{SO}(3)$ denote the global hand pose and the local rotation of joint $j$, respectively, at time step $t$. We represent frame-to-frame motion as local pose increments obtained by applying the logarithm map $\operatorname{Log}(\cdot)$, using the same notation for both $\text{SE}(3)$ and $\text{SO}(3)$ for brevity, to consecutive relative transformations:
\begin{equation}
    \boldsymbol{\xi}_t = \operatorname{Log}(\boldsymbol{T}_t^{-1}\boldsymbol{T}_{t+1}),\ \boldsymbol{\omega}_{t,j} = \operatorname{Log}(\boldsymbol{R}_{t,j}^{-1}\boldsymbol{R}_{t+1,j}),
\end{equation}
where $\boldsymbol{\xi}_t\in\mathbb{R}^{6}$ comprises translational and rotational components, and $\boldsymbol{\omega}_{t,j}\in\mathbb{R}^{3}$ represents the local joint rotation increment. The temporal regularization is formulated as
\begin{equation}
\begin{aligned}
    \mathcal{L}_{\text{temp}} = \frac{1}{\Delta t^5}\sum_t \Bigl(&\lambda_{\text{lin}}\left\|\Delta^2\boldsymbol{\xi}_t^{\text{trans}}\right\|_2^2+\lambda_{\text{ang}}\left\|\Delta^2\boldsymbol{\xi}_t^{\text{rot}}\right\|_2^2 \\
    +&\lambda_{\text{joint}}\sum_j\left\|\Delta^2\boldsymbol{\omega}_{t,j}\right\|_2^2\Bigr).
\end{aligned}
\end{equation}
where $\Delta^2\boldsymbol{x}_t=\boldsymbol{x}_{t+2}-2\boldsymbol{x}_{t+1}+\boldsymbol{x}_t$ denotes the second-order finite difference and $\Delta t$ is the time interval between consecutive frames. This formulation penalizes third-order temporal variations in hand motion while respecting the geometry of rigid-body and joint rotations.

To evaluate the effect of temporal refinement, we compare the mean inter-frame 3D joint displacement before and after optimization. Specifically, given the world-space positions $\boldsymbol{x}_{t,j}\in\mathbb{R}^3$ of the $J$ hand joints at time step $t$, we define
\begin{equation}
    d_t = \frac{1}{J}\sum_{j=1}^{J} \left\|\boldsymbol{x}_{t,j} - \boldsymbol{x}_{t-1,j}\right\|_2,
\end{equation}
which measures the average joint displacement between consecutive frames, capturing both global hand motion and finger articulation. As illustrated in \cref{app:fig:hand_refinement}, the initial reconstructions exhibit abrupt displacement spikes, particularly during rapid in-hand manipulation. Temporal refinement substantially suppresses these fluctuations, yielding more temporally consistent hand trajectories.

\subsection{Object Trajectory Refinement}

For object pose estimation, we optimize the sequence of 6D object poses by balancing consistency with the initial estimates against temporal regularization:
\begin{equation}
    \mathcal{L}_{\text{obj}}=\lambda_{\text{pose}}\mathcal{L}_{\text{pose}}+\lambda_{\text{temp}}\mathcal{L}_{\text{temp}}.
\end{equation}
The pose consistency term $\mathcal{L}_{\text{pose}}$ employs the Geman--McClure robust loss to penalize translational and symmetry-aware rotational discrepancies between the refined and initial object poses, reducing the influence of unreliable pose estimates. The temporal term $\mathcal{L}_{\text{temp}}$ penalizes linear and angular jerk to encourage smooth object trajectories.

Object symmetries pose an additional challenge for temporal refinement, as visually equivalent configurations may yield inconsistent orientation estimates, resulting in abrupt discontinuities in the estimated trajectories. To address this issue, we explicitly account for discrete and continuous rotational symmetries when evaluating pose consistency. Let $\mathcal{G}$ denote the rotational symmetry group of an object. The symmetry-aware rotational discrepancy between an estimated orientation $\boldsymbol{R}$ and a reference orientation $\tilde{\boldsymbol{R}}$ is defined as
\begin{equation}
    d_{\mathcal{G}}(\boldsymbol{R},\tilde{\boldsymbol{R}}) = \min_{\boldsymbol{S}\in\mathcal{G}}\left\|\operatorname{Log}\left(\boldsymbol{R}^{\mathsf{T}}\tilde{\boldsymbol{R}}\boldsymbol{S}\right)\right\|_2.
\end{equation}
This formulation measures rotational discrepancies modulo object symmetries, avoiding unnecessary penalties between symmetry-equivalent orientations.

As illustrated in \cref{app:fig:symmetry}, the initial pose estimates exhibit abrupt orientation changes caused by symmetry ambiguities, including 180$^\circ$ axis flips during pen spinning and 90$^\circ$ rotational ambiguities during prism manipulation. By resolving these ambiguities and improving temporal consistency, the refinement produces smoother object trajectories without introducing artificial rotational jumps.

\section{Additional Qualitative Examples}

We provide additional example sequences from \dataset to further illustrate the diversity of high-speed dexterous hand-object interactions captured in \dataset. 

\begin{figure*}[!t]
    \centering
    \includegraphics[width=\linewidth]{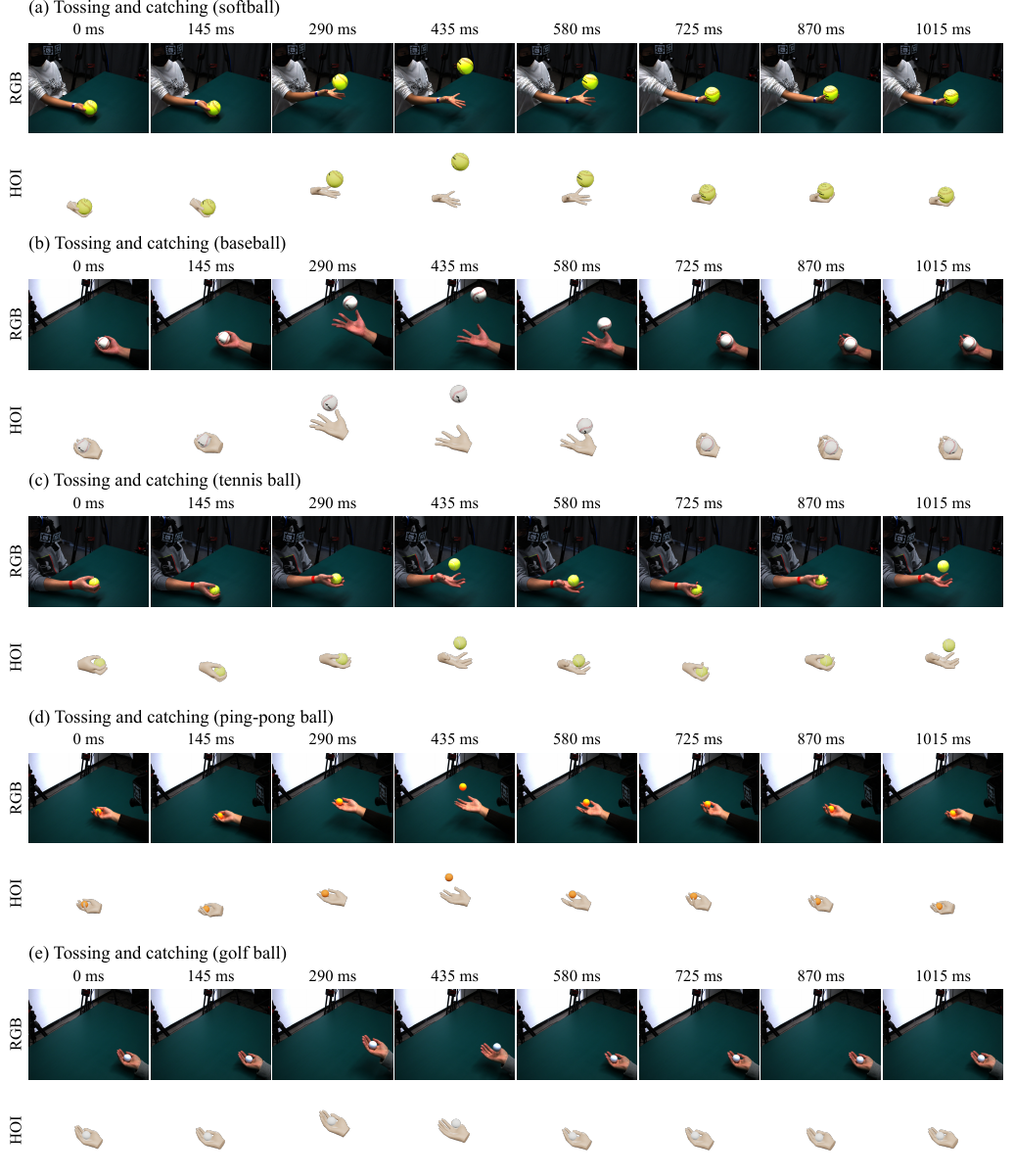}
    \caption{Additional qualitative examples from \dataset (1/3).}
\end{figure*}

\begin{figure*}[!t]
    \centering
    \includegraphics[width=\linewidth]{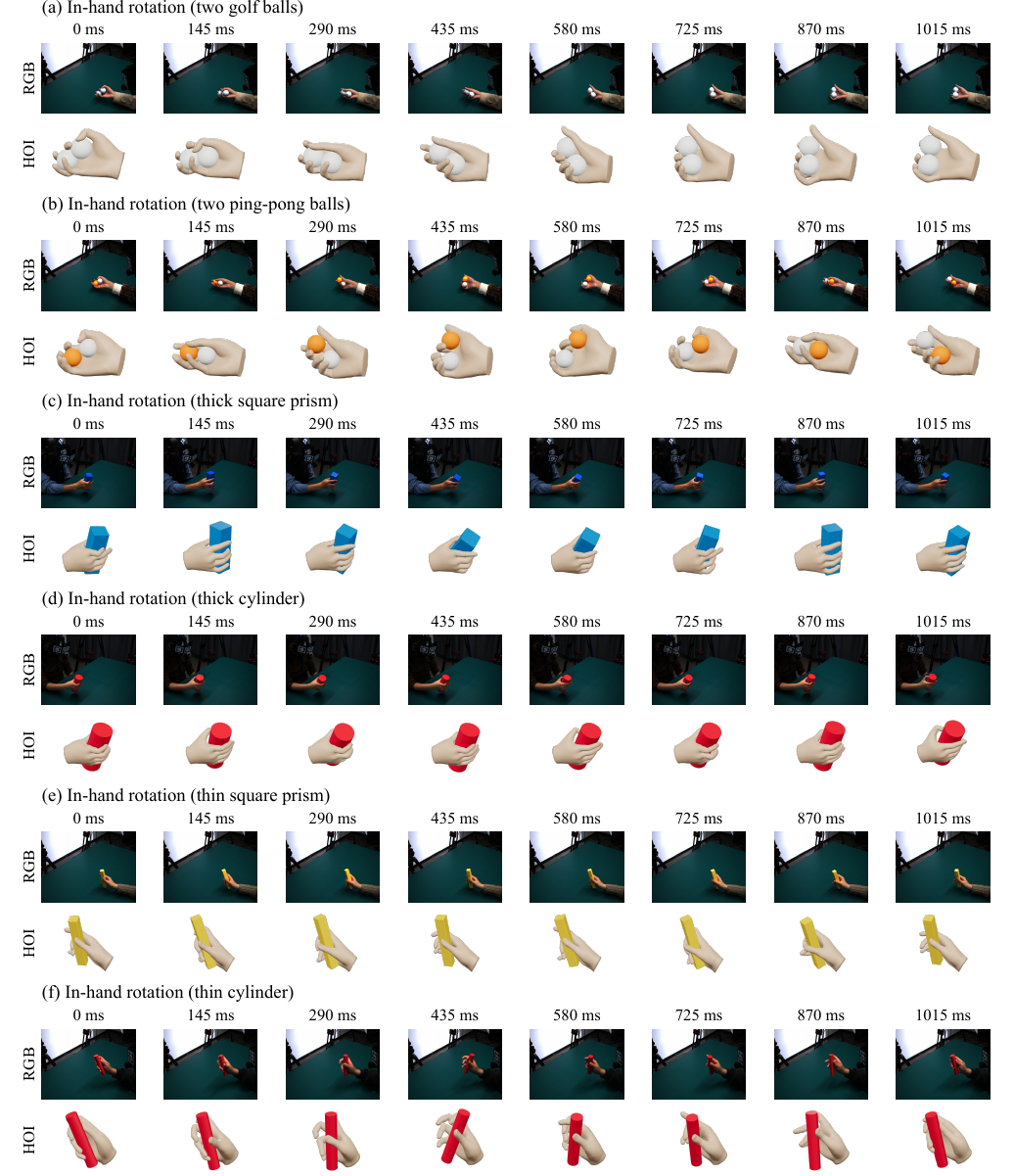}
    \caption{Additional qualitative examples from \dataset (2/3).}
\end{figure*}

\begin{figure*}[!t]
    \centering
    \includegraphics[width=\linewidth]{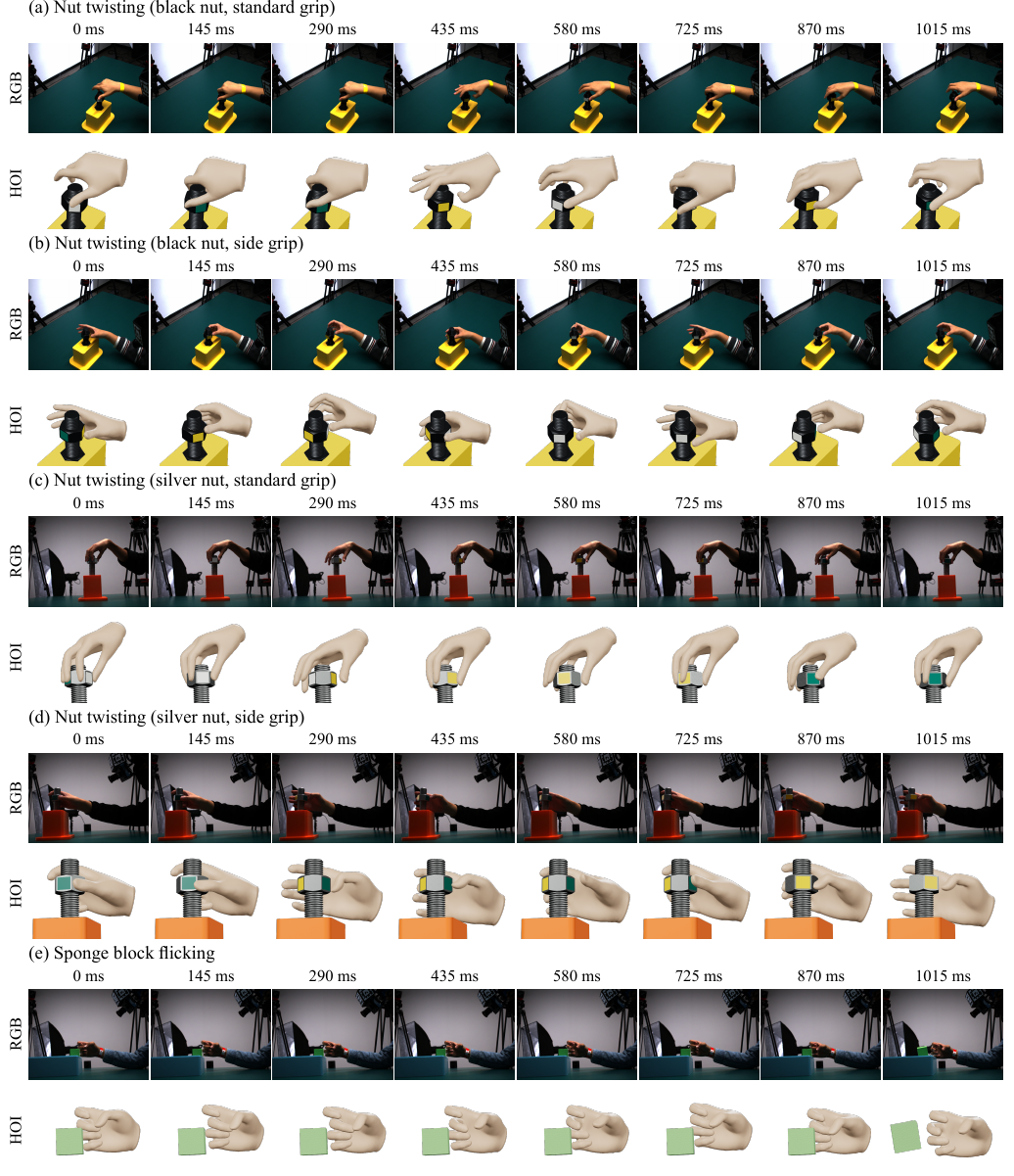}
    \caption{Additional qualitative examples from \dataset (3/3).}
\end{figure*}

\end{document}